\PassOptionsToPackage{sort&compress}{natbib}
\documentclass[letterpaper, 10 pt, conference]{ieeeconf} 

\let\labelindent\relax

\IEEEoverridecommandlockouts                              
\usepackage{graphicx}	
\usepackage{amsmath}	
\usepackage{amssymb}	
\usepackage{booktabs}
\usepackage{times}
\usepackage{microtype}
\usepackage{epsfig}
\usepackage[table,xcdraw,dvipsnames]{xcolor}
\usepackage{arydshln}
\usepackage{placeins}
\usepackage{color, colortbl}
\usepackage{stfloats}
\usepackage{enumitem}
\usepackage{tabularx}
\usepackage{xstring}
\usepackage{multirow}
\usepackage{xspace}

\usepackage{graphicx}
\usepackage{amsmath}
\usepackage{amssymb}
\usepackage{booktabs}
\usepackage{array}
\usepackage{multirow}
\usepackage{makecell}
\usepackage{bbding}
\PassOptionsToPackage{normalem}{ulem}
\usepackage{ulem}
\usepackage{arydshln}
\usepackage{xcolor}         
\usepackage[accsupp]{axessibility}
\usepackage{pifont}
\usepackage{graphicx}
\usepackage{amssymb}
\usepackage{xcolor}
\usepackage{pifont}
\usepackage{multirow}
\usepackage{authblk}

\usepackage[colorlinks=true, urlcolor=cyan, filecolor=green, citecolor=blue]{hyperref}

\title{\LARGE \bf
Force-EvT: A Closer Look at Robotic Gripper Force Measurement with \\ Event-based Vision Transformer
}

\begin{document}

\author{Qianyu Guo$^{1}$, Ziqing Yu$^{2}$, Jiaming Fu$^{3}$, Yawen Lu$^{4}$, Yahya Zweiri$^{5}$, Dongming Gan$^{6}$
 
 \thanks{$^{1}$Qianyu Guo is a Ph.D. student in the School of Engineering Technology at Purdue University, USA.
         {\tt\small guo716@purdue.edu}}%
 \thanks{$^{2}$Ziqing Yu is a Ph.D. student in the School of Engineering Technology at Purdue University, USA.
         {\tt\small yu1154@purdue.edu}}%
 \thanks{$^{3}$Jiaming Fu is a Ph.D. student in the School of Engineering Technology at Purdue University, USA.
         {\tt\small fu330@purdue.edu}}%
 \thanks{$^{4}$Yawen Lu is with the Computer Graphics Technology Department at Purdue University, USA.
         {\tt\small lu976@purdue.edu}}%
 \thanks{$^{5}$Dr. Yahya Zweiri is a Professor of Aerospace Engineering at Khalifa University, UAE.
         {\tt\small yahya.zweiri@ku.ac.ae}}%
 \thanks{$^{6}$Dr. Dongming Gan is an Associate Professor in the School of Engineering Technology at Purdue University, USA.
         {\tt\small dgan@purdue.edu}}
 }

\maketitle

\thispagestyle{empty}
\pagestyle{empty}
%

\begin{abstract}

Robotic grippers are receiving increasing attention in various industries as essential components of robots for interacting and manipulating objects. While significant progress has been made in the past, conventional rigid grippers still have limitations in handling irregular objects and can damage fragile objects. We have shown that soft grippers offer deformability to adapt to a variety of object shapes and maximize object protection. At the same time, dynamic vision sensors (e.g., event-based cameras) are capable of capturing small changes in brightness and streaming them asynchronously as events, unlike RGB cameras, which do not perform well in low-light and fast-moving environments.
In this paper, a dynamic-vision-based algorithm is proposed to measure the force applied to the gripper. In particular, we first set up a DVXplorer Lite series event camera to capture twenty-five sets of event data. Second, motivated by the impressive performance of the Vision Transformer (ViT) algorithm in dense image prediction tasks, we propose a new approach that demonstrates the potential for real-time force estimation and meets the requirements of real-world scenarios. 
We extensively evaluate the proposed algorithm
on a wide range of scenarios and settings, and show that it consistently outperforms recent approaches.

\end{abstract}


\section{INTRODUCTION}


The robotic hand represents a critical component of a robot, typically mounted on the robot's arms. A key part of the robotic hand is the gripper, which facilitates interaction with the environment and manipulation of target items. Gripping mechanisms find extensive application across various industrial sectors, including the food industry ~\cite{lien2013gripper}, healthcare~\cite{kyrarini2021survey}, and agriculture~\cite{dhanawade2018review}. Traditional grippers often employ rigid metals as their primary material. However, such rigid grippers lack flexibility in handling objects with irregular shapes and may inadvertently damage items made from fragile materials. In contrast, innovative soft robotic grippers can deform to accommodate the shape and size of the target object, thereby enhancing the object's protection. Consequently, soft grippers emerge as the preferred option for many manipulation tasks~\cite{fu2023variable, fu2023machine}.

\begin{figure*}[htbp]
	\center{\includegraphics[width=0.93\linewidth, height=7.7cm]{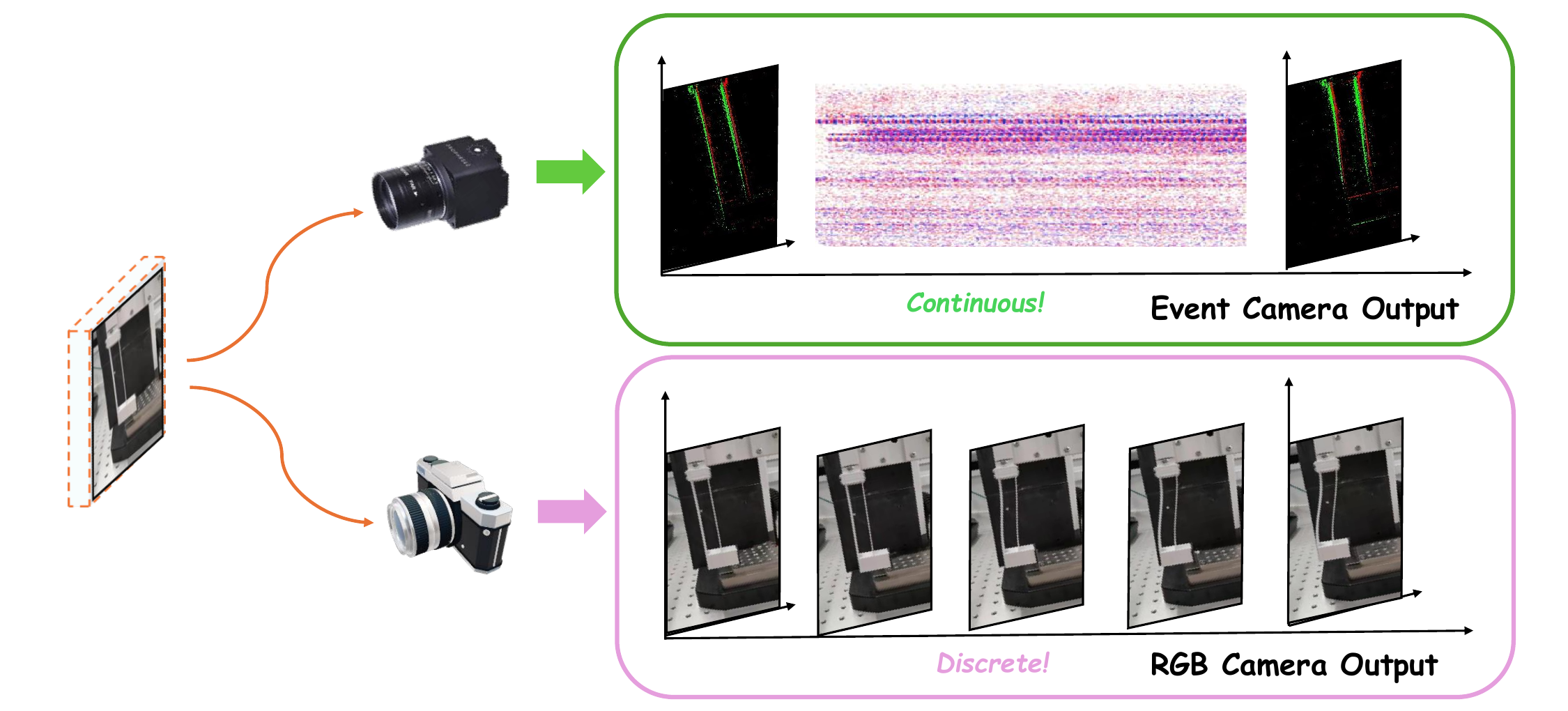}} 
    \vspace{0.5mm}
	\caption{Comparison of frame-based RGB Camera and event camera outputs in experimental scenarios, reveals significant distinctions in how these two types of cameras capture and process visual information.}
	\label{event camera}
\end{figure*}

Using vision-based methods to predict the deformation and force applied to the robotic gripper is a popular topic, while the use of traditional RGB cameras is always the first choice. However, when the experimental environment is dark or the gripper is moving very fast, RGB cameras cannot clearly capture the trajectory of objects, or even get images with significant motion blur, where thermal imaging~\cite{lu2021superthermal} and LiDAR~\cite{lu2024lidar} are often utilized as compensation. Event cameras, known as bio-inspired sensors, are able to detect very slight brightness changes at the pixel-level and output them as events, as shown in Fig. \ref{event camera}. The events include four important information: timestamps, x coordinates, y coordinates, and their polarities. Compared to standard cameras, event cameras have four remarkable advantages. \textit{High Temporal Resolution}: Capture slight brightness changes fast, and output events in the order of $\mu$s. \textit{Low Power Consumption}: Due to their efficient design, which only transmits brightness changes and does not output redundant data, event cameras achieve low power consumption. \textit{Wide Dynamic Range}: Event cameras can acquire visual information by over 120 dB, exceeding standard cameras by over 60 dB, thanks to logarithmic-scale photoreceptors and asynchronous pixel operation. \textit{Low Latency}: Event cameras have ultra-low latency since pixels detect and transmit changes independently without global exposure timing \cite{gallego2020event}. Therefore, event cameras have a strong ability to capture gripper motion, even if the experimental environment is not perfect.

The cooperation of machine learning and event cameras is a new way of solving computer vision problems, which achieves great performance \cite{scheerlinck2020fast, muthusamy2020neuromorphic, naeini2019novel}. For example, object detection \cite{perot2020learning}, object tracking \cite{ramesh2018long}, 3D reconstruction \cite{rebecq2018emvs}, steering prediction for self-driving cars \cite{maqueda2018event}, optical flow and intensity estimation \cite{bardow2016simultaneous}, etc. These methods typically use a continuous stream of asynchronous events, which allows for efficient processing. Nevertheless, due to the sparse and unstructured nature of the event streams, it is a challenge to directly observe and process the event data \cite{weng2021event}. To better adapt to the traditional frame-based computer vision algorithms, most event data is converted into event frames based on timestamp or polarity.

\begin{figure*}[htbp]
	\center{\includegraphics[width=0.92\linewidth, height=10.5cm]{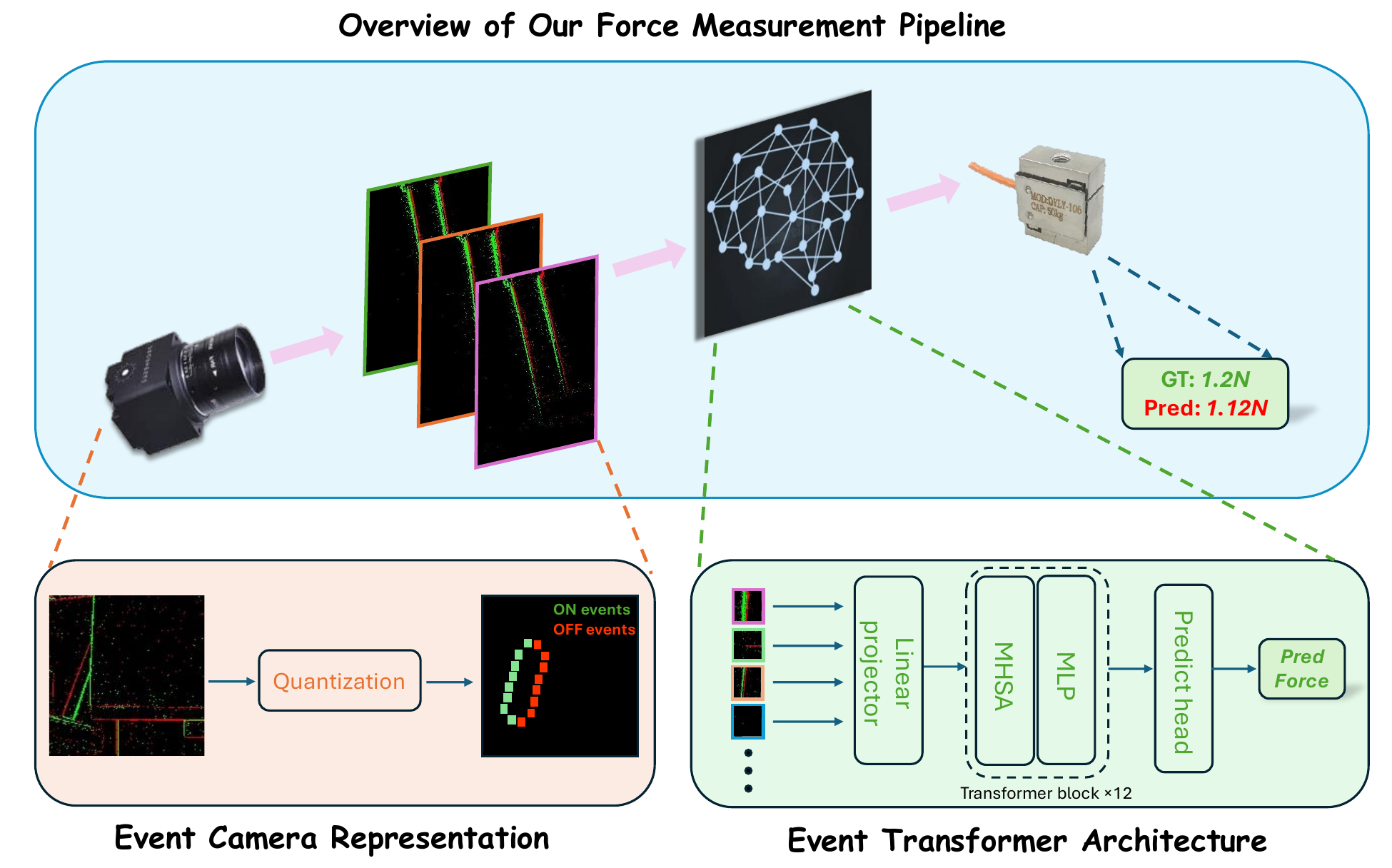}} 
    \vspace{0.5mm}
	\caption{Overview of the proposed method Force-EvT. The output of the event camera is converted into event frames over a certain time interval $T$. Then, the event frames are processed by Vision Transformer (ViT) network, which makes force prediction of the robotic gripper.}
	\label{framework}
\end{figure*}

The advantage of exploring the Vision Transformer on force measurement via event frames has become evident. Notably, in a real-world scenario (Fig. \ref{framework}), we achieve a 13.0\% percentage error and 0.13 N RMSE compared to the ground truth force sensor measurement, benefiting from the \textit{event frame representation} and \textit{event transformer architecture}.
In the application of our previous proposed variable stiffness robotic gripper \cite{fu2023variable}, it is an important part of outputting real-time required force to the control end. ArUco marker detection strategy was used to monitor the deformation of the gripper, enabling the control end to read the force applied to the grippers efficiently. However, while applying this technique in a less-than-ideal environment, for example, if the illumination is not enough, the marker detection would become intermittent. Therefore, in the paper, we propose a vision-based Vision Transformer in force measurement task, which is able to capture the slight deformation of the robotic gripper and make force predictions with high accuracy.

The main contributions of this work are summarised as:

\begin{itemize}
\item We propose a novel approach to estimate the force applied to a robotic gripper using a Dynamic Vision Sensor.

\item We collect a dataset using our custom-designed robotic gripper, a force sensor and an event camera, named RG-Event, which contains 1000 event frames and their corresponding force labels.

\item We utilize state-of-the-art Vision Transformer architecture as a backbone to train the data collected using an event camera. We show that Vision Transformer performs well in regression tasks.

\end{itemize}

\section{Related Work}

In this section, we explore various recent advancements in event-based vision and Vision Transformer techniques, as well as existing methodologies for force measurement in robotics, encompassing traditional contact-type force sensors, and sensorless approaches.

\subsection{Event-based Vision}

Event-based vision is a developing technique and has great potential. Due to their advantages of high temporal resolution, low power consumption, low Latency and high dynamic range, these bio-inspired visual sensors are able to be used in various complex environments. In \cite{maqueda2018event}, the authors proposed a deep neural network for steering angle prediction using event cameras, showcasing superior performance in challenging conditions like low light and fast motion compared to traditional cameras. Weng et al. first presented a novel Transformer-based network called ET-Net for event-based video reconstruction \cite{weng2021event}. GSCEventMOD was an approach for detecting moving objects based on events, which had great performance in challenging scenarios such as fast movements and sudden changes in lighting conditions \cite{mondal2021moving}. In addition, a lot of computer vision applications such as optical flow estimation, depth estimation, motion segmentation, and visual-inertial odometry all achieved excellent performances using event-based methods \cite{gallego2020event}.

\subsection{Vision Transformer in Prediction}
Transformer refers to a type of neural network architecture that was initially utilized in natural language processing (NLP) tasks, such as machine translation and text generation \cite{dosovitskiy2020image}. Inspired by the successful utilization in NLP, Transformer has been gradually applied to computer vision tasks \cite{wang2021pyramid,sun2021rethinking,chen2022vision}, which largely improved conventional CNN and LSTM based networks \cite{weimin2024enhancing,dai2023addressing}. For instance, Ranftl et al. presented a novel architecture called dense prediction transformers, which employed Vision Transformer instead of convolutional networks as the foundational structure for tasks requiring dense predictions \cite{ranftl2021vision}. In \cite{ramana2023vision}, Vision Transformer was employed alongside Convolutional Neural Networks (CNN) to forecast urban traffic congestion. TransDepth is also a novel transformer-based network, aiming to make pixel-wise predictions in various computer vision tasks, such as depth estimation, surface normal estimation. Lu et al. introduced TransFlow which used a pure transformer for optical flow estimation, demonstrating the effectiveness of incorporating spatial self-attention and cross-attention mechanisms \cite{lu2023transflow}.

\subsection{Force Measurement}
The existing force measurement methods applied to robot hands usually fall into two categories, namely traditional force sensors and non-sensor methods. Researchers have proposed various force sensors integrated into gripper structures. In \cite{kawasaki2002dexterous}, the Gifu hand II is introduced, featuring the capability to be equipped with a six-axis force sensor at each fingertip, showcasing high integration levels. Dai et al. developed a contact force transducer based on a six-component Stewart platform to enhance reliability and precision \cite{dai2000six}. In \cite{kuang2017design}, Kuang et al. introduce a novel hinged-joint cantilever beam sensor structure designed to reduce sensor nonlinearity. The aforementioned traditional contact-type force sensors typically exhibit high accuracy and reliability but come with drawbacks, such as occupying substantial structural space and increasing the complexity of the structure. Therefore, with the advancements in technologies like machine learning and computer vision, researchers have interdisciplinarily proposed new sensorless methods for force sensing. For instance, \cite{xu2021compliant} captures deformations in nodes located on the framework of the fin ray gripper, \cite{zhang2018robot} observes the movement of markers on the soft layer, \cite{zhu2019flexure} measures changes in the angle of the fingers. Furthermore, Baghaei Naeini et al. proposed a dynamic-vision based approach to measure contact force on silicone membrane, using Convolutional Recurrent Neural Networks \cite{baghaei2020dynamic}. These sensorless methods simplify the structure of robotic hands. However, they also face the challenge of insufficient precision. Additionally, the RGB cameras used to capture deformations operate continuously, potentially consuming plenty of system resources and resulting in high energy consumption.

\section{Methodology}

As shown in Figure \ref{framework}, our methodology is designed to precisely quantify the forces applied to a robotic gripper, utilizing data captured by an event camera. In this section, we first address the conversion of raw event data into a structured frame format, according to a certain time interval. Then, we employ a regression algorithm based on the Vision Transformer architecture to estimate the forces applied to the robotic gripper. Finally, we introduce the loss function that guides training towards precise predictions.

\subsection{Event Frame based Representation} Deep learning algorithms, which stand at the forefront of recent advancements in machine learning, have been developed with a focus on processing conventional frame-based data. To bridge the gap between the unique data structure produced by event cameras and the requirements of these advanced algorithms, we first perform an event-to-frame conversion. In this case, the asynchronous events will be converted into synchronous frames. As mentioned before, an event, as captured by an event camera, consists of four key attributes:  the x-coordinate ($x_{i}$), the y-coordinate ($y_{i}$), the timestamp ($t_{i}$), and the polarity ($p_{i}$) of the change in brightness. Therefore, we use $e_{i} = (x_{i}, y_{i}, t_{i}, p_{i})$ to represent the attributes. 

In general, event-to-frame conversion can be approached through various methodologies, primarily based on timestamps, the number of events, or the polarity of events \cite{zheng2023deep}. In this work, we adopt a timestamp-based approach for constructing event frames. Given a certain time interval \textit{T}, a flow of events would be divided into numerous event-based frames. Since the event camera is highly sensitive to changes of brightness, the slight deformation of the robotic gripper could be clearly captured, and the edges of the gripper are clearly depicted in each event frame. 

\begin{figure}[t!]
\begin{center}
\includegraphics[width=0.95\linewidth,height=58mm]{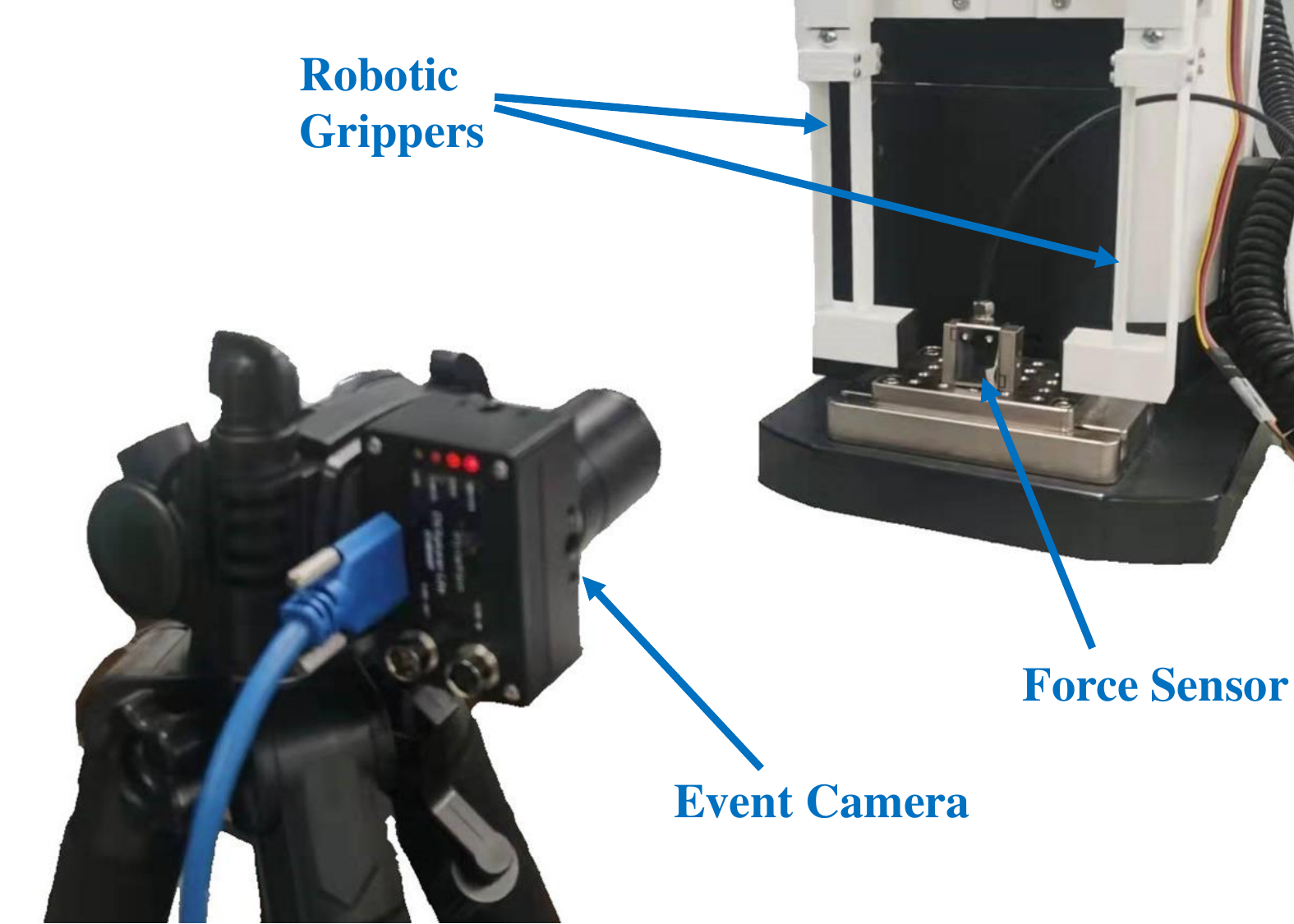} 
\end{center}
\vspace{0.5mm}
   \caption{Experimental setup to conduct data collection using an event camera, a force sensor, and robotic grippers.
  }
\label{setup}
\end{figure}

\subsection{Vision Transformer based Force Measurement}
Vision Transformer (ViT) is a powerful deep learning architecture that can be used in computer vision tasks. In this work, we leverage the ViT as a foundational architecture to estimate the forces applied to a robotic gripper, an application where precision and contextual understanding of spatial relationships are important. 

Unlike traditional Convolutional Neural Networks (CNNs) which analyze images through a hierarchical sequence of localized filters, ViT approaches the task by dividing the input event frames into fixed-size patches (8 $\times$ 8 pixels in our implementation). Each patch is then transformed into a high-dimensional vector through a linear embedding process. This transformation not only preserves the spatial hierarchy of the original image but also allows for a more granular analysis of the visual content. Once embedded, these patches are fed into a series of transformer encoder, allowing the model to capture both local and global features within patches and learn the relationships across the entire image. For force estimation on a robotic gripper, this means the ViT can intelligently focus on critical regions of the input frames that are most indicative of the applied forces, such as areas of significant deformation or contact points between the gripper and objects. In addition, the adaptability and efficiency of ViT are further enhanced by its self-attention mechanism, which allows for selective focus on salient features within the image patches, effectively ignoring irrelevant information \cite{dosovitskiy2020image}.

\subsection{Loss Function}
In the development of our force measurement model, an essential component is the choice of an appropriate loss function to guide the training process towards accurate predictions. For this purpose, we employ the Mean Squared Error (MSE) as loss function in our experiments \cite{sara2019image, guo2021multi}. MSE is widely recognized for its efficacy in regression tasks and its ability to quantify the variance between predicted values and ground truth. The selection of MSE is motivated by its sensitivity to large errors, making it particularly suitable for ensuring precision in force measurement.

The MSE is mathematically defined as the average of the squared differences between the predicted forces ($\hat{y}_{i}$) and the actual measured forces (${y}_{i}$), overall $N$ samples in the dataset. The formula for MSE is given by:

\begin{equation}
MSE = \frac{1}{N}  {\textstyle \sum_{N}^{i=1}} (\hat{y}_{i}  - {y}_{i})^{2} 
\end{equation}

During the model training phase, the minimization of MSE facilitates the adjustment of model parameters that incrementally improves the accuracy of force predictions.

\section{Experiments}
In this section, we first present the experimental settings of the proposed approach. Following this, we introduce the process of data collection and data preprocessing steps taken to prepare the data for analysis. Finally, force measurement evaluations are provided.

\begin{figure}
\begin{center}
\includegraphics[width=0.99\linewidth,height=45mm]{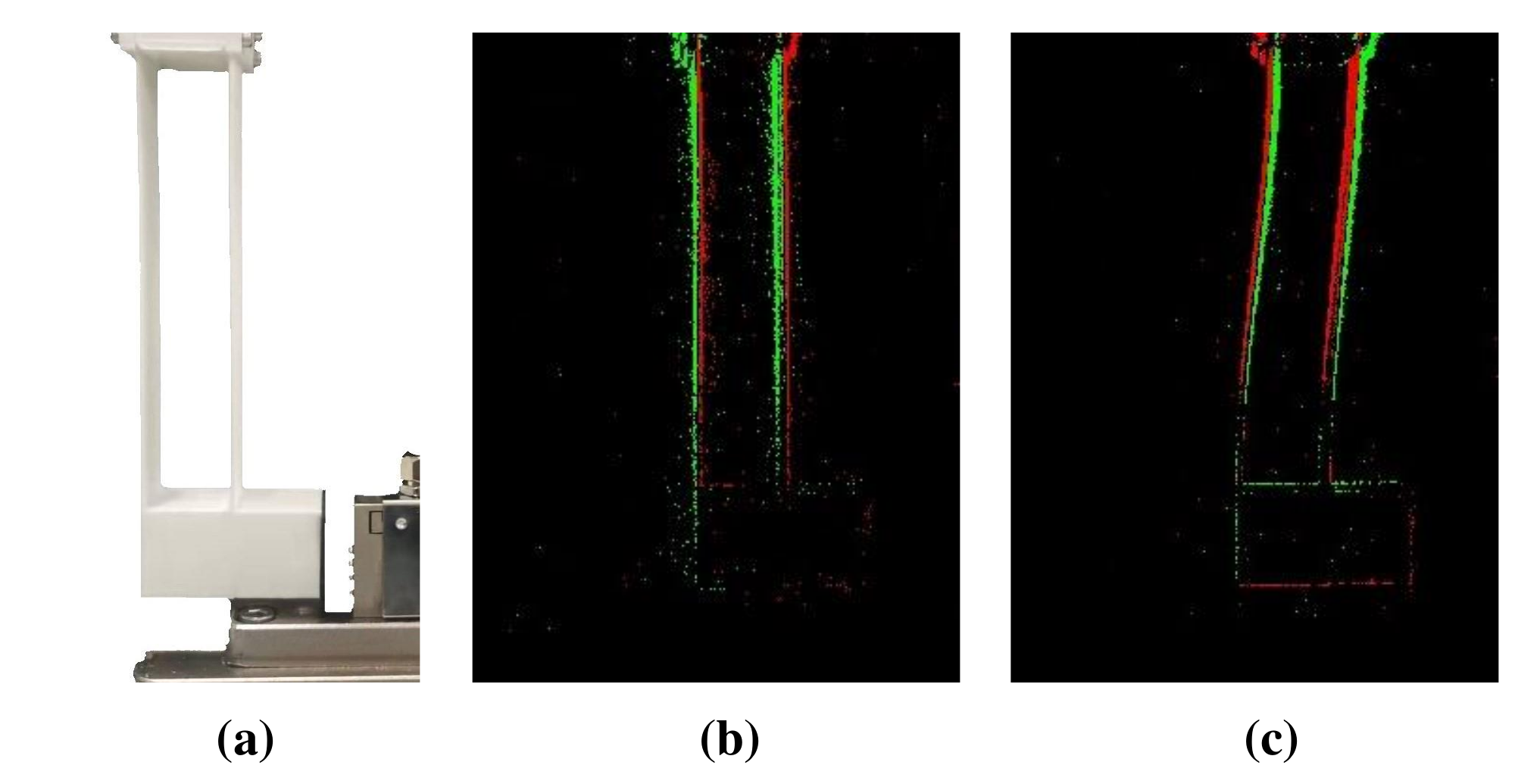} 
\end{center}
\vspace{0.5mm}
   \caption{The gripper is captured by an RGB camera and an event camera. (a) displays our designed soft robotic gripper captured by a standard RGB camera in a state without any applied force, (b) and (c) show the gripper under different deformation states, as captured by an event camera. 
  }
\label{dataset}
\end{figure}

\subsection{Experimental Setup} 
As shown in Figure \ref{setup}, in this work, the experimental setup contains a tension and compression sensor (commonly referred to as a force sensor), an Arduino microcontroller, an event camera (DVS sensor), and a custom 3D-printed robotic gripper. The event camera, known for its high-speed and low-latency imaging capabilities, provided asynchronous visual information crucial for dynamic scenes. The force sensor, grasped by the robotic gripper, enabled precise measurement of the forces exerted during object manipulation tasks. An Arduino microcontroller served as the central processing unit, orchestrating data acquisition and communication between the force sensor and robotic gripper. In the experiments, the force sensor is positioned at the center of the two grippers to ensure optimal force measurement. Through the action of grasping the robotic grippers, the force sensor receives the applied force data at a rate of 10 samples per second. 

In the implementation, we train and test the specific vit\_base\_patch8\_224 model on our collected dataset. In the experiment, the dataset is randomly divided into a training set, a validation set, and a testing set in the ratio of 7: 1.5: 1.5. Our force prediction training uses an Adam optimizer with a learning rate set to 0.001 and a batch size of 16. This entire pipeline is deployed on two GeForce RTX 3090s GPU platforms using the PyTorch framework.

\subsection{Dataset}
To estimate the force exerted on our designed soft robotic gripper, we conduct a comprehensive data collection using the event camera. Employing the experimental setup described previously, we repeat the process of closing, grasping, and opening the gripper for 25 times. Given our interest in capturing the critical moments when the gripper undergoes deformation, our dataset exclusively encompasses event frames corresponding to the grasping phase. 

Finally, a total of 1000 event frames were gathered using the DVXplorer Lite camera, and we call the dataset as \textbf {RG-Event}. During each experimental iteration, the force applied to the gripper varied within a range from 0 N to 1.6 N throughout the grasping stage. We synchronize the data collection windows for both the force sensor and the event camera to a duration of $T$ = 100 ms, ensuring precise temporal alignment between the sensory inputs and the visual data. Illustrated in Figure \ref{dataset}, there are two representative images extracted from our dataset, showcasing the progressive deformation of the gripper from its initial state to the point of maximum deformation.

\begin{figure}[t!]
\begin{center}
\includegraphics[width=0.85\linewidth,height=52mm]{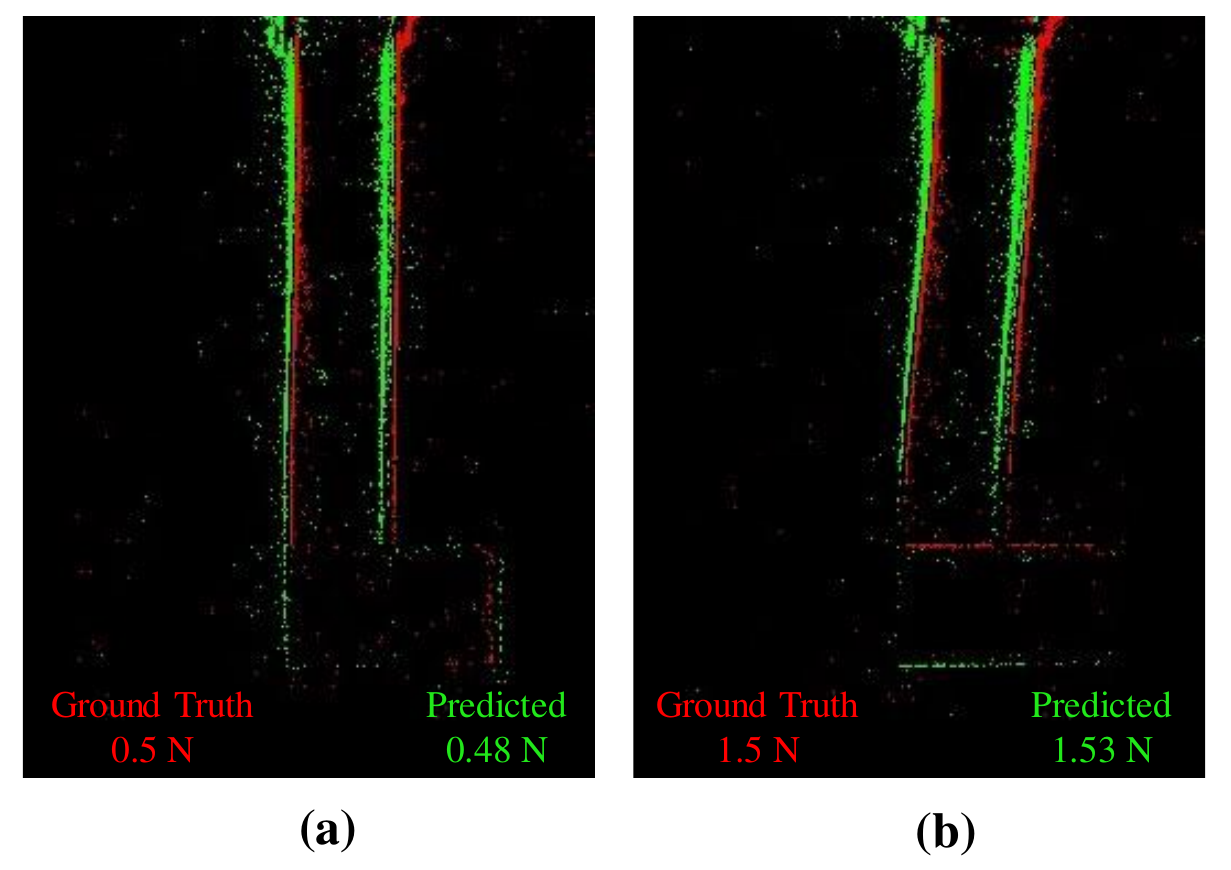} 
\end{center}
\vspace{0.5mm}
   \caption{The prediction results demonstrate the efficacy and accuracy of employing Force-EvT in force measurement task.
  }
\label{Evaluation}
\end{figure}

\subsection{Force Measurement Evaluations}

In order to evaluate the performance of using ViT in the force prediction task, we use the Root Mean Squared Error (RMSE) \cite{chai2014root, wen2021distortion} and R-squared ($R^2$) \cite{cameron1997r} as evaluation metrics. RMSE is a measure of the average deviation of the predictions made by a model from the actual observed values. Lower RMSE values indicate better performance of the model, as it means the model's predictions are closer to the actual values. $R^2$ is a statistical measure that represents the proportion of the variance in the dependent variable that is predictable from the independent variables in a regression model. It ranges from 0 to 1, where $R^2$ value closer to 1 indicates a better fit of the model to the data. In our testing stage, we get RMSE as 0.13 N and $R^2$ as 0.93. As shown in Figure \ref{Evaluation}, the deformed grippers with 0.5 N and 1.5 N are predicted as 0.48 N and 1.53 N respectively. The prediction results with high accuracy demonstrate the effectiveness of the proposed method. Furthermore, as shown in Table \ref{table1}, we provide the performance comparisons between the novel event-based approach and our previous marker-based approach \cite{fu2023variable}.

\begin{table}[]
\centering
\scalebox{1.1}{
\renewcommand\arraystretch{1.35}
\begin{tabular}{c|c|c|c}
                      & $RMSE$ & $R^2$ & $Error$  \\ \hline
Marker-based Approach \cite{fu2023variable} & -    & -                    & 19.5\% \\ \hline
Force-EvT      & 0.13 & 0.93                 & 13\%  
\end{tabular}}
\vspace{0.5mm}
\caption{Our novel Force-EvT model is able to achieve better performance in force measurement compared with our previous marker-based method.}
  \label{table1}
\end{table}

\section{Conclusion and Future Work}

In this paper, we introduce a novel approach named Force-EvT for predicting forces applied to soft robotic grippers using event-based vision. Leveraging a Dynamic Vision Sensor, particularly the DVXplorer Lite event camera, we capture and record gripper deformation processes. By employing the Vision Transformer (ViT) algorithm, our proposed method demonstrates promising results and potential for force estimation in robotic applications. Experimental evaluations validate the effectiveness of the approach, highlighting its suitability for measuring forces applied to soft robotic grippers.

For future works, we intend to expand our experiments to encompass different illumination conditions, including both very bright and dark environments, to demonstrate the superiority of using event cameras in force measurement projects. Moreover, we plan to incorporate more complex designs of robotic grippers into our training data. By diversifying our dataset, we can enhance the robustness and adaptability of our approach to a wider range of gripper configurations and applications.

\section*{ACKNOWLEDGMENT}

This work is supported by the National Science Foundation (NSF) grant under CMMI-2131711.



\bibliographystyle{abbrv-doi}
\bibliography{template}

\end{document}